\documentclass[letterpaper]{article} 
\usepackage{aaai25}  
\usepackage{times}  
\usepackage{helvet}  
\usepackage{courier}  
\usepackage[hyphens]{url}  
\usepackage{graphicx} 
\usepackage{natbib}  
\usepackage{caption} 
\usepackage{algorithm}
\usepackage{algorithmic}
\usepackage{amsmath}

\usepackage{newfloat}
\usepackage{listings}
\DeclareCaptionStyle{ruled}{labelfont=normalfont,labelsep=colon,strut=off} 
\floatstyle{ruled}
\newfloat{listing}{tb}{lst}{}
\floatname{listing}{Listing}
\title{Enhancing Close-up Novel View Synthesis via Pseudo-labeling}

\author{
    Jiatong Xia\equalcontrib,~ Libo Sun\equalcontrib\thanks{},~ Lingqiao Liu\\
}
\affiliations{
    Australian Institute for Machine Learning, The University of Adelaide\\

    \{jiatong.xia,~libo.sun,~lingqiao.liu\}@adelaide.edu.au
}

\usepackage{bibentry}

\begin{document}


\twocolumn[{%
\renewcommand\twocolumn[1][]{#1}%
\maketitle
    \captionsetup{type=figure}
    \centering
    \includegraphics[width=0.88\textwidth]{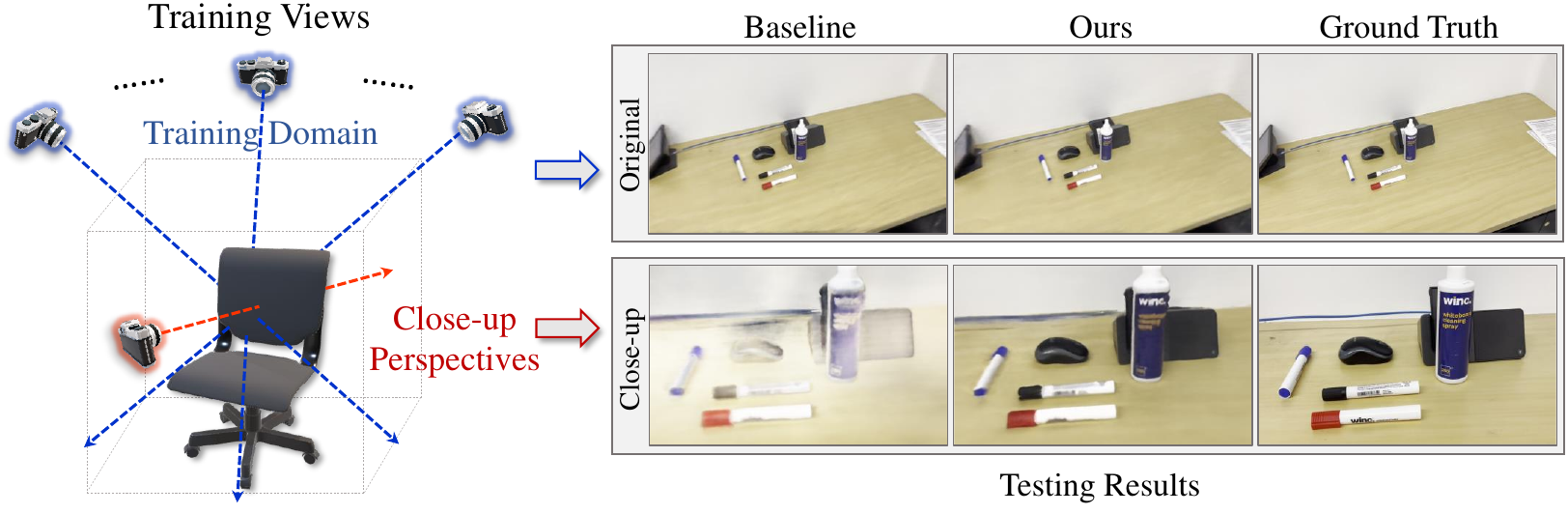} 
    \hfill\caption{This work investigate novel view synthesis from viewpoints significantly different from those in the training data (blue camera), particularly focus on diverse close-up perspectives (red camera). When facing such close-up perspectives, radiance field methods often exhibit artifacts similar to those seen in the baseline results depicted above. We initially delve into this particular issue, and our proposed method in this paper significantly enhanced radiance fields in such situation.}
    \label{fig:teaser}
    \hfill
    \vspace{2mm}
}]
{
\renewcommand{\thefootnote}{\fnsymbol{footnote}}
\footnotetext[1]{These authors contributed equally.}
\footnotetext[2]{Corresponding author.\\Copyright © 2025, Association for the Advancement of Artificial
Intelligence (www.aaai.org). All rights reserved.}
}

\begin{abstract}
Recent methods, such as Neural Radiance Fields (NeRF) and 3D Gaussian Splatting (3DGS), have demonstrated remarkable capabilities in novel view synthesis. However, despite their success in producing high-quality images for viewpoints similar to those seen during training, they struggle when generating detailed images from viewpoints that significantly deviate from the training set, particularly in close-up views. The primary challenge stems from the lack of specific training data for close-up views, leading to the inability of current methods to render these views accurately.
To address this issue, we introduce a novel pseudo-label-based learning strategy. This approach leverages pseudo-labels derived from existing training data to provide targeted supervision across a wide range of close-up viewpoints. Recognizing the absence of benchmarks for this specific challenge, we also present a new dataset designed to assess the effectiveness of both current and future methods in this area. Our extensive experiments demonstrate the efficacy of our approach. The code is at: \url{https://github.com/JiatongXia/Pseudo-Labeling.git}
\end{abstract}

\section{Introduction}
In recent years, there has been a surge in the use of radiance field approaches, such as Neural Radiance Fields (NeRF)~\cite{mildenhall2020nerf}, for view synthesis. To extend the practical applications of radiance fields, various enhancements have been introduced, including methods aimed at increasing processing efficiency~\cite{chen2021mvsnerf, neff2021donerf, yu2021plenoctrees, kurz2022adanerf} and enabling image manipulation capabilities~\cite{lin2021anycost, zhang2021pise, wang2023seal, kuang2023palettenerf}. However, despite the extensive development of radiance field techniques, these methods often fall short when tasked with producing high-quality images from viewpoints that significantly differ from those seen in the training data. A particularly challenging case is the generation of close-up views, which are often sought by users who wish to examine the fine details of an object from specific angles, particularly in situations where it's difficult to position a camera at the desired location in real-world scenarios.

In this paper, we investigate the use of radiance fields to generate close-up views from model trained on distant viewpoints. This involves training on images that capture scenes from afar, while the goal is to produce detailed close-up views of specific elements from various angles. The primary challenge lies in the lack of diverse close-up views within the training dataset, which leaves the radiance field unoptimized for rendering accurate close-up perspectives.

To address this challenge, we propose refining the training protocol for radiance fields by integrating pseudo-annotations derived from a carefully crafted pseudo-labeling approach. Our method involves generating a virtual close-up viewpoint at random during each training iteration. For each virtual viewpoint, we create wrapped images by mapping pixels from the original training images based on the rendered depth. These wrapped images are then evaluated for consistency and occlusion to determine which pixels can be effectively used as pseudo-training data. These pseudo-training data are subsequently incorporated into the general training process to enhance performance for close-up views.

Additionally, recognizing that users may prefer to achieve high-quality results for specific close-up views by slightly fine-tuning an existing model rather than training a new model, we introduce a test-time fine-tuning method that significantly improves performance for specific close-up views while requiring minimal processing time.
Furthermore, we have developed a dataset specifically designed to evaluate the generation of close-up views. This dataset addresses the current lack of benchmarks for assessing the performance of existing and future methods in this domain.

\section{Related Work}
\subsubsection{Neural Radiance Fields.} The original NeRF, introduced by Mildenhall et al.~\cite{mildenhall2020nerf}, represents a scene as a continuous 5D function that maps spatial coordinates and viewing directions to radiance values. Since its introduction, NeRF-related techniques have found applications in various computer vision tasks~\cite{zhang2021ners, chen2022tensorf, azinovic2022neural, liu2024one, ssdnerf}.
In the context of enhancing human-machine interaction, several works~\cite{wang2022clip, chen2022tensorf, bao2023sine} have demonstrated NeRF's capabilities. Liu et al.~\cite{liu2021editing} proposed a method for NeRF editing, introducing a technique to propagate coarse 2D user scribbles into 3D space for shape and color modification. Yuan et al.~\cite{yuan2022nerf} developed a method that enables controllable shape deformation within the implicit scene representation, allowing for scene editing without network re-training. Kerr et al.~\cite{kerr2023lerf} incorporated raw CLIP embeddings into NeRF to support diverse natural language queries across real-world scenes. 
In the context of limited data, DS-NeRF~\cite{deng2022depth} leverages sparse depth data to provide additional supervision, thus improving the performance of radiance fields.
Mip-NeRF~\cite{barron2021mip} efficiently rendering anti-aliased conical frustums instead of rays, reduces aliasing artifacts and improves NeRF's capability to capture fine details.
To accelerate processing speed, Instant-NGP~\cite{mueller2022instant} uses a hash grid and an occupancy grid to accelerate computation and a smaller MLP to represent density and appearance to achieve faster training.
Zip-NeRF~\cite{barron2023zipnerf} further integrates advancements from scale-aware anti-aliased NeRFs and fast grid-based NeRF training, combining their strengths to enhance performance and efficiency.

\subsubsection{Gaussian Splatting.} 
Alongside the widespread application of NeRF, 3D Gaussian Splatting (3DGS)~\cite{kerbl3Dgaussians} has emerged as a powerful technique for novel view synthesis. Compared to NeRF, 3DGS requires less training time and enables high-quality, real-time novel view synthesis at high resolution. Yuan et al.~\cite{Huang2DGS2024} further proposed a novel approach, 2D Gaussian Splatting (2DGS), which incorporates depth distortion and normal consistency to more accurately model and reconstruct geometrically precise radiance fields.
More recently, Yu et al.~\cite{Yu2024MipSplatting} introduced Mip-Splatting, which employs two sampling filters to limit the maximum frequency of Gaussian primitives and approximate the box filter, thereby simulating the physical imaging process. 
While their approach is innovative in addressing aliasing issues, it diverges from our focus as it does not effectively handle viewpoints that significantly differ from those in the training domain (e.g., close-up views with varying camera orientations).

\section{Method}
\subsection{Preliminary:  Neural Radiance Fields} 
We would use neural radiance fields (NeRF) as a representative radiance field approach to introduce our method. NeRF utilize an MLP network to map a 3D location $\mathbf{x} \in {R}^{3}$ and a viewing direction $\mathbf{d} \in {R}^{3}$ to color values $\mathbf{c} \in {R}^{3}$ and a volume density $\mathbf{\sigma}$. The mapping function $\mathcal{F}_{\theta}$ can be defined as $\mathcal{F}_{\theta}(\mathbf{x}, \mathbf{d}) = (\mathbf{c}, \mathbf{\sigma})$, where $\theta$ represents the learnable parameters of the MLP.
To generate a pixel in a novel view, a camera ray $\mathbf{r}(t) = \mathbf{o} + t\mathbf{d}$ is back-projected from the camera center $\mathbf{o}$ in the direction of $\mathbf{d}$. 
The color of a pixel is then rendered as: 
\begin{equation}
\label{eq_c}
\hat{\mathbf{C}}(\mathbf{r} )=\sum_{i=1}^{N}w_{i}\mathbf{c}_{i}~,
\end{equation}
\begin{equation}
\label{eq_w}
\mathrm{with} ~~~~ w_{i}=T_{i}(1-\mathrm{exp} (-\sigma _{i}\delta _{i}))~,
\end{equation}
\begin{equation}
~~~~~~~~~~~~T_{i}=\mathrm{exp} \left ( - \sum_{j=1}^{i-1} \sigma _{j} \delta _{j}   \right ) ~,
\end{equation}
\begin{equation}
\delta_{i}=t_{i+1}-t_{i}~.~~~~~
\end{equation}
where $t_i \in [t_{n}, t_{f}]$, with $t_{n}$ representing the near bound and $t_{f}$ representing the far bound of a ray.
To optimize the network parameters $\theta$, an RGB MSE loss is imposed between the rendered pixels and the pixels of training images:
\begin{equation}
\label{eq_loss}
\mathcal{L}_{c}(\mathbf{r} )=\left\| \hat{\mathbf{C} }(\mathbf{r} ) - \mathbf{C}(\mathbf{r} ) \right\|_{2}^{2}~.
\end{equation}
Previous methods only provide supervision within the training rays sample. As a result, their performance in viewpoints significantly different from those in the training data becomes uncertain, which leads to the problem that we aim to address.

\subsection{The Problem of Close-up Observations}
After optimizing the radiance field through the process outlined in the previous section, any ray with camera center $\mathbf{o}$ and ray direction $\mathbf{d}$ similar to the training rays can be sampled through the MLP network to obtain RGB results that closely approximate the quality of the training ground truth after rendering.
This process can be conceptually understood as utilizing the optimized radiance field to interpolate the unknown rays within the training ground truth data.

However, as depicted in Fig.~\ref{fig:teaser}, when we move a camera pose closer and simultaneously randomly alter its orientation, the resulting rays will have camera centers $\mathbf{o}$ and ray directions $\mathbf{d}$ that significantly diverge from those in the training data.
These rays often yield unreliable results in the final rendering output, typically manifesting as artifacts.
Specifically, NeRF network is configured with a branch to output RGB values. At the initial stage of this branch, the direction of the ray $\mathbf{d}$ is taken as input. when an untrained ray direction is input for a sampling point $\mathbf{x}$, the network tends to produce an unreliable RGB value $\mathbf{c}$. Such sampling points often lead to artifacts in the final volume rendering (Eq.~\ref{eq_c}).

In terms of density, the radiance field itself exhibits robustness in restoring geometric information (Eq.~\ref{eq_w}), therefore, the density results of sample points on rays in the untrained domain are typically robust than RGB results. Nonetheless, as we bring the camera closer, due to the absence of constraints on rays in the this distance and direction, minor density deviations in original training domain are amplified.

\begin{figure*}[!t]
\centering
\includegraphics[width=0.74\linewidth]{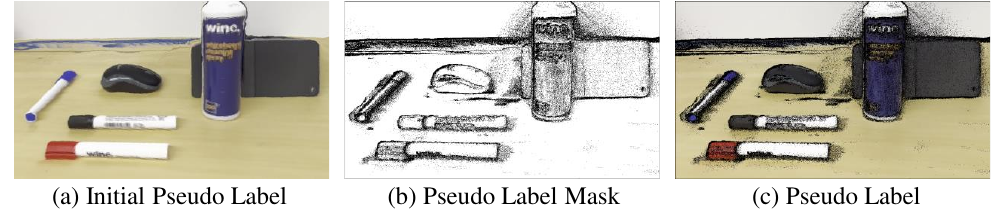} 
\caption{The generated pseudo labels. We show the intermediate outputs in the pseudo label generation process: (a) The initial pseudo label extracted from training images (i.e., $\mathbf{I}'_n$), 
(b) The pseudo label mask as described in Eq.~\ref{eq_mask}, and  
(c) The final pseudo-label obtained after applying the mask.}
\label{fig:pseudolabel}
\end{figure*}

\subsection{Pseudo-labeling for Diverse Close-up Perspectives}
Drawing from the knowledge presented above, inference rays outside the training samples exhibit erroneous density and color due to insufficient learning from training data. This deficiency leads to errors such as artifacts when rendering close-up observations across diverse perspectives.
To tackle this issue, our goal is to generate reliable training labels for perspectives that involve close-up observations and use these generated labels to train and enhance the radiance field, thereby improving the performance of close-up observations.
To begin with, we propose an approach to generate camera poses for diverse close-up perspectives. Then, we demonstrate how to generate reliable labels for each generated camera pose.

\subsubsection{Diverse Close-up Perspectives Generation.} 
To efficiently fine-tune the radiance field in close-up perspectives, it is imperative to generate diverse perspectives outside the training perspectives which have images captured from far distances. 
Specifically, each camera pose is generated through computations involving the geometry of the radiance field and the training perspectives $\left \{ \mathbf{P}_{1},\mathbf{P}_{2}, ...\mathbf{P}_{N}  \right \}$.
Initially, for a random selected training pose $\mathbf{P}_{n}$,
\begin{equation}
    \mathbf{P}_{n} = [\mathbf{R}_{n} | \mathbf{t}_{n}]~~{\rm with}~~ n\in(0,N),
\end{equation}
its corresponding depth map $\mathbf{D}_{n}$ can be rendered from the radiance field.
Then, a pixel $(u_{a}, v_{a})$ in $\mathbf{D}_{n}$ is randomly selected as the anchor point for generating a close-up perspective. For the pixel $(u_{a}, v_{a})$, its corresponding 3D position $\mathbf{X}_{a}=(x_{a}, y_{a}, z_{a})$ in the world coordinate can be obtained as:
\begin{equation}
\mathbf{X}_{a} = \mathbf{o}_{a} + \mathbf{D}_{n}(u_{a},v_{a})\cdot\mathbf{d}_{a},
\end{equation}
where the camera center $\mathbf{o}_{a}=\mathbf{t}_{n}$ and $\mathbf{d}_{a}$ is the ray direction calculated from $\mathbf{R}_{n}$. 

To generate a close-up perspective, we randomly select a 3D point between the camera center of $\mathbf{P}_{n}$ and point $\mathbf{X}_{a}$ to form a new camera position $\mathbf{t}'_{n}$:
\begin{equation}
\label{eq_t}
    \mathbf{t}'_{n} = \frac{\left( \left (\lambda -1\right )\mathbf{\cdot X}_{a} + \mathbf{t}_{n}   \right)}{\lambda },
\end{equation}
where $\lambda$ is the magnification of how closer the camera is to the anchor point.

To obtain the sample space of camera rotations, we use the rotation matrix $\mathbf{R}_{n}$ of the original training pose as the reference. The conversion between Euler angles and rotation matrices is a common operation in 3D perception, and we define these two conversion as $F_{\textbf{R}\Rightarrow \textbf{e}}$ and $F_{\textbf{e}\Rightarrow \textbf{R}}$. 
Therefore, $\mathbf{R}_{n}$ can be converted into Euler angles as:
\begin{equation}
F_{\textbf{R}\Rightarrow \textbf{e}}(\mathbf{R}_{n}) = (\theta_{x_{n}},\theta_{y_{n}},\theta_{z_{n}}),
\end{equation}
where $(\theta_{x_{n}},\theta_{y_{n}},\theta_{z_{n}})$ represent the values of Euler angles.
After converting $\mathbf{R}_{n}$ to Euler angles, 
the new camera orientation $\mathbf{e'}$ is generated as:
\begin{equation}
\label{eq_e}
\mathbf{e}'_{n} = \left\{\begin{matrix}
\theta_{x}' ~=~ \theta_{x_{n}} + ~\Delta \theta_{x} ~, ~~\Delta \theta_{x}\in (\theta_{x_{n}}- \varepsilon,~\theta_{x_{n}}+\varepsilon)
\\
\theta_{y}'~ =~ \theta_{y_{n}} + ~\Delta \theta_{y} ~, ~~\Delta \theta_{y}\in (\theta_{y_{n}}- \varepsilon,~\theta_{y_{n}}+\varepsilon)
\\
\theta_{z}'~ = ~\theta_{z_{n}} + ~\Delta \theta_{z} ~, ~~\Delta \theta_{z}\in (\theta_{z_{n}}- \varepsilon,~\theta_{z_{n}}+\varepsilon),
\end{matrix}\right.    
\end{equation}
where $\Delta \theta_{x}$, $\Delta \theta_{y}$, and $\Delta \theta_{z}$ are three randomly generated offsets, and $\varepsilon$ is a parameter used to ensure that the camera orientation does not change too drastically.
Once the new camera orientation is obtained, its corresponding rotation matrix $\mathbf{R}'_n$ can be obtained as:
\begin{equation}
   \mathbf{R'}_n = F_{\textbf{e}\Rightarrow \textbf{R}}(\mathbf{e}'_{n}).
\end{equation}
Finally, the camera pose for a randomly generated close-up perspective is written as:
\begin{equation}
    \mathbf{P}'_{n} =[\mathbf{R}'_{n} | \mathbf{t'} ].
\end{equation}

\subsubsection{Pseudo Labels and Masks.}
With a pre-trained radiance field (i.e. a NeRF model), we can render the depth of each pixel in a given view. For a virtual pose $\mathbf{P'}_n$ from the camera pose $\mathbf{P}_n$, we can render its depth map $\mathbf{D}'_n$ and thus calculate the 3D coordinates of each pixel. Then we can find projections those coordinates on a training view $\mathbf{I}_n$ and their corresponding pixel value. This could lead to a wrapped image $\mathbf{I}'_n$ by extracting pixel values from $\mathbf{I}_n$ based on the depth estimation $\mathbf{D}'_n$. This process can be denoted as $\mathbf{I'}_{n}  =  W\left(\mathbf{I}_{n}, \mathbf{D}'_n,\mathbf{P}'_n, \mathbf{P}_{n}, \mathbf{K}  \right)$, where $\mathbf{K}$ is the camera intrinsic matrix.

Alternatively, we can render the depth of each pixel in $\mathbf{I}_n$ and find their correspond projections in $\mathbf{I}'_n$. One can also copy the pixel value from $\mathbf{I}_n$ to the corresponded pixel in $\mathbf{I}'_n$ to create a wrapped image $\mathbf{I}^*_n$. It important to note that not every pixel in $\mathbf{I}_n$ can find correspondence in view $\mathbf{I}'_n$, the wrapped image should contain some undefined pixels.  This process can be denoted as $\mathbf{I^*}_{n}  =  W\left(\mathbf{I}_{n}, \mathbf{D}_n,\mathbf{P}'_n, \mathbf{P}_{n}, \mathbf{K}  \right)$.

Ideally, if the depth estimation is accurate, those two wrapped images should have same RGB values at the corresponding pixels. However, due to the error of depth estimation in $\textbf{D}$ and $\textbf{D}'$, those values may not be consistent. If their pixel value difference is small enough, we will reasonably assume that their corresponding depth estimation is correct.

The wrapped image $\mathbf{I}^*_n$ for pose $\mathbf{P}'_n$ can be obtained from multiple training view images $\{\mathbf{I}_{1}, \mathbf{I}_{2}, ..., \mathbf{I}_{N}\}$, each resulting in a wrapped image. In other words, for a given pixel $(u,v)$, there might be $\mathbf{I}^*_n(u,v)$, $n \in \{1,\cdots,N\}$, $N$ possible values. We aggregate them by always retaining the pixel value corresponding to the minimum depth. This is because occlusions need to be considered, where nearby points can block distant points. 
By warping all the training view images and checking for occlusions, we can update $\mathbf{I}^*_n$ to $\mathbf{\bar{I}}^*_n$, which contains projected pixels from all the training view images while considering occlusions.

The consistency between $\mathbf{I}^{'}_{n}$ and $\mathbf{\bar{I}}^*_n$ can be checked to remove errors contained in $\mathbf{I}^{'}_{n}$ for composing the pseudo label used in training.
For each pixel $(u, v)$ in image $\mathbf{I}^{'}_{n}$, the mask $\mathbf{M}_{n}$ which decides if $\mathbf{I}^{'}_{n}(u, v)$ will be used as a pseudo label for fine-tuning is defined as:
\begin{equation}
\label{eq_mask}
\textbf{M}_{n}(u, v) = \begin{cases}
 \rm True & \text{ if } ~~|\mathbf{I}^{'}_{n}(u, v) -  \mathbf{\bar{I}}^*_n(u, v)| < \epsilon   \\
 \rm False & \text{ otherwise, }
\end{cases}  
\end{equation}
 where $\epsilon$ denotes the threshold for determining whether two RGB values can be considered as matched, and we set $\epsilon$ to 0.05 in this paper. An example of a generated pseudo-label is shown in Fig.~\ref{fig:pseudolabel}. As we can observe from this figure, our method can generate accurate pseudo-labels and remove areas that contain errors.

\begin{figure*}[!t]
\centering
\includegraphics[width=0.79\linewidth]{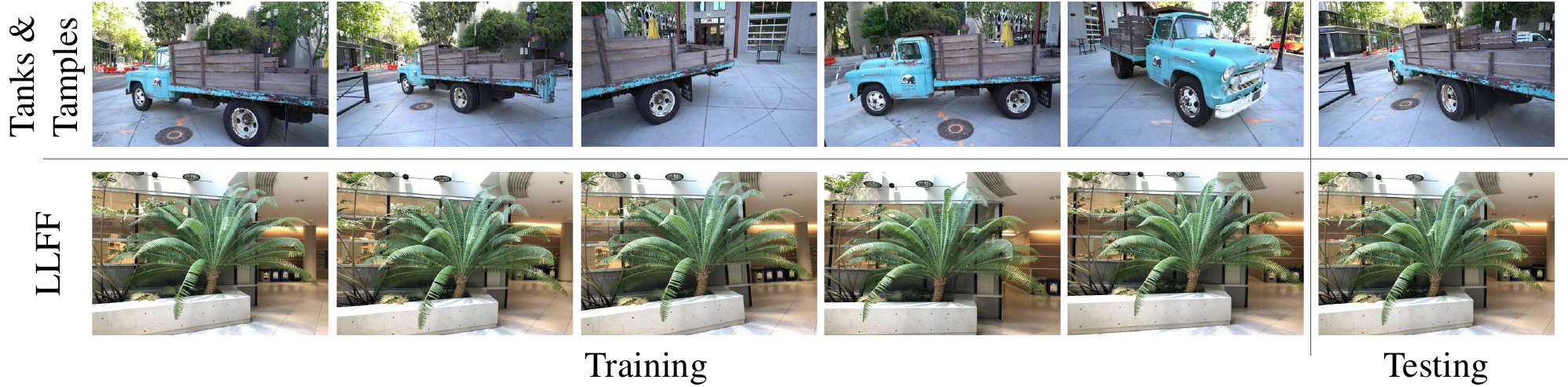} 
\caption{
Typical existing view synthesis benchmarks. The test images are positioned at the similar distance as the training images and share highly similar view directions. }
\label{fig:other_benchmark}
\end{figure*}
\begin{figure*}[!t]
\centering
\includegraphics[width=0.80\linewidth]{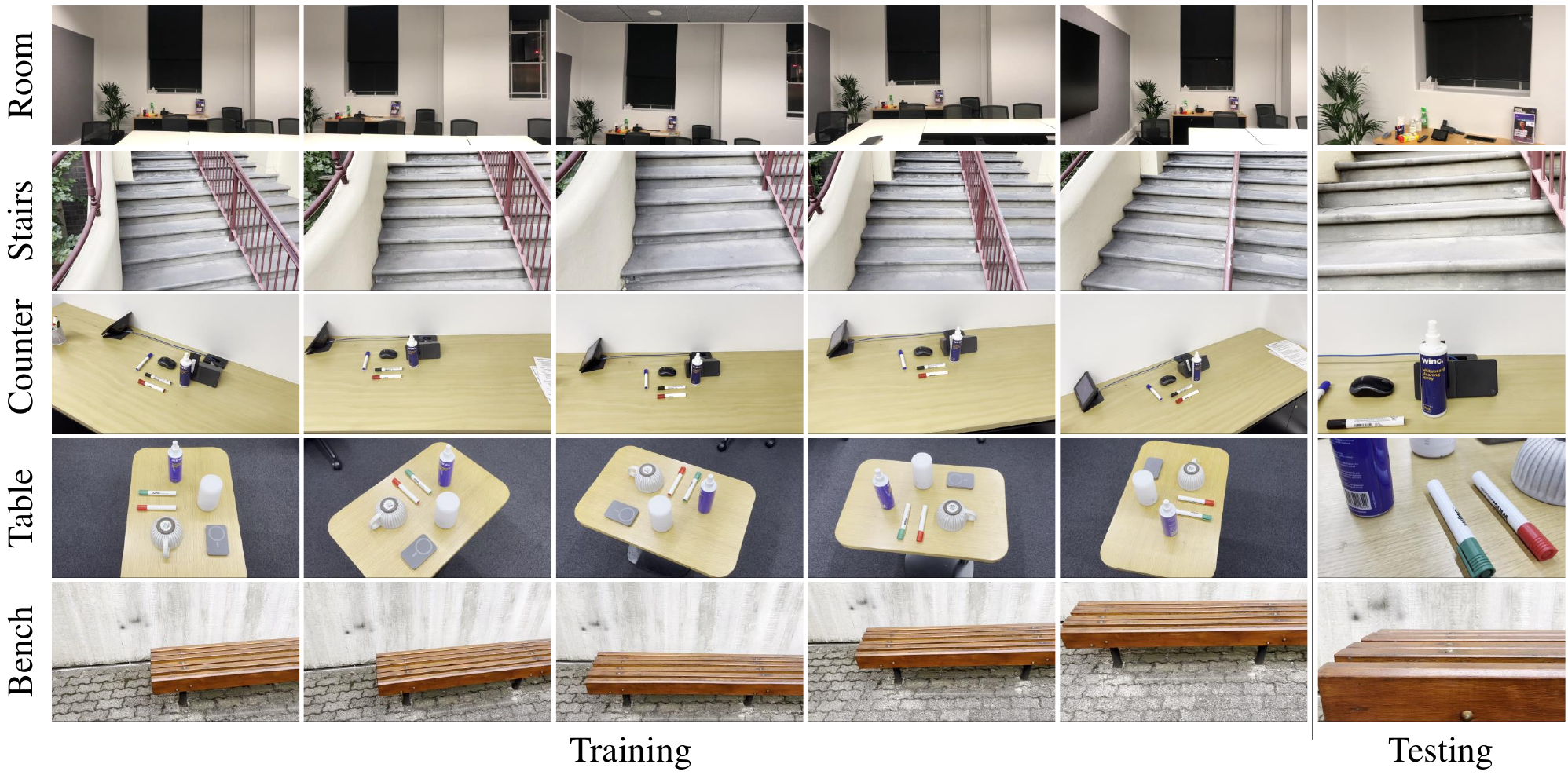} 
\caption{
Our dataset with each row as the example of a scene, where the training images on the left and the testing images on the right. Training images in each scene are in the same domain of the example, moving forward facing or moving around the objects in a similar distance, while testing images are much closer to the objects and are significantly divergent from the training views.}
\label{fig:benchmark}
\end{figure*}

\subsection{Training on Diverse Close-up Perspectives}
In order to include rays from a wide variety of diverse perspectives during training, we adopt a strategy of generating a new virtual camera pose along with its corresponding pseudo labels at each fine-tuning iteration. 
We randomly select the value of $\lambda$ in Eq.~\ref{eq_t} from the range of (2, 8) for each iteration. This implies that the virtual camera pose in each iteration can be randomly positioned closer from $\frac{1}{2}$ to $\frac{1}{8}$ of the distance between the original training camera location and the anchor point. And set the value of $\varepsilon$ in Eq.~\ref{eq_e} to $\frac{\pi }{4}$.
We set the fine-tuning process for 10k iterations, resulting in a total of 10k random virtual camera poses used for fine-tuning, significantly covers the untrained regions.

\subsubsection{Batchify Random Rays for NeRF.}
When fine-tuning NeRF, rendering the entire depth map for each virtual pose at every iteration can be extremely time-consuming. Therefore, we propose a batchify random virtual perspectives training.

In each training iteration, instead of rendering the entire depth map for $\mathbf{P'}_n$, we randomly choose a set of 2D coordinates $\left \{ (u_{1},v_{1}), (u_{2},v_{2}),...,(u_{B},v_{B})  \right \}$, where $B$ represent the batch size.
Using these 2D coordinates, we can sample a training batch-size collection of rays $\left \{ \mathbf{r}_{1},\mathbf{r}_{2},...,\mathbf{r}_{B} \right \}$ from virtual camera pose $\mathbf{P'}_n$, then render the depth values $\left \{ z_{1},z_{2},..., z_{B} \right \}$ for this set of rays through radiance field and derive pseudo labels $\left \{ \mathbf{I'}_{n}(u_{k},v_{k}  )  \right \}_{k=1}^{B}$ for this set of rays. 

We combine this batch of pseudo labels with the same batch size of original training samples, then fine-tune the radiation field to encompass diverse perspectives across both trained and untrained domains.
The supervision provided by pseudo labels here can be represented as:
\begin{equation}
      \mathcal{L}_{pl}=\left\| \hat{\mathbf{C}} - \hat{\mathbf{I'}_{n}} \right\|_{2}^{2}.
\end{equation}
And the overall loss function $\mathcal{L}$ can be expressed as follows:
\begin{equation}
\label{eq_loss_all}
    \mathcal{L} = \frac{1}{2}\left ( \mathcal{L}_{c} + \mathcal{L}_{pl} \right ).
\end{equation}

\subsubsection{Fine-tuning on Gaussian Splatting.}
Instead of per pixel rendering, the tile-based fast rasterizer of Gaussian Splatting could render an entire image and its depth map at a time. This allows us to directly produce the entire pseudo label $\mathbf{I'}_{n}$ and its corresponding mask $\mathbf{M}_{n}$ for generated close-up pose $\mathbf{P'}_n$,
thus we can apply $\mathbf{I'}_{n}$ to provide supervision for Gaussian Splatting fine-tuning after filtering with $\mathbf{M}_{n}$. We choose 2DGS~\cite{Huang2DGS2024} as our baseline method of Gaussian Splatting in this paper due to its robust capability to generate reliable depth maps for each synthesized view.

\subsection{Test-time Fine-tuning} 
In practical scenarios, users may occasionally have specific perspectives in mind that they wish to use. For instance, they might want to observe a cup on a table from particular viewpoints. In such cases, fine-tuning the radiance field on these predetermined perspectives can significantly reduce the iteration times required for fine-tuning with randomly generated camera poses and accelerate the process. 

We consider test-time fine-tuning as optional and as an additional benefit. 
Specifically, creating virtual camera poses in an extensive untrained domain and refining the radiance field by applying pseudo labels with masks can enhance the overall rendering quality for diverse perspectives. And directly generating pseudo-labels and masks for the camera poses that need to be tested in the untrained domain, and applying a `test-time fine-tuning' on these testing poses, is more targeted and can lead to rapid convergence, while this setup aligns perfectly with the Gaussian Splatting methods due to their quick training time and the ability for the tile-based rasterizer to render an entire view during optimization.
Typically, only 5 iterations are needed to complete the fine-tuning on one specific test view (less than 3 seconds on our test images).

Unlike the generated training perspectives, where imaged are chosen based on a randomly selected anchor to obtain pseudo labels, we introduce a selection strategy to determine which image will be used to obtain pseudo labels. We first project the pixels of all training images $\{\mathbf{I}_{1}, \mathbf{I}_{2}, ..., \mathbf{I}_{N}\}$ onto this specific test pose $\mathbf{P''}_n$ using their respective training depths $\{\mathbf{D}_{1}, \mathbf{D}_{2}, ..., \mathbf{D}_{N}\}$, and count the number of pixels from each training image that can be projected onto this test pose, recorded as $\{\mathbf{\tau }_{1}, \mathbf{\tau }_{2}, ..., \mathbf{\tau }_{N}\}$.
By assessing $\{\mathbf{\tau }_{1}, \mathbf{\tau }_{2}, ..., \mathbf{\tau }_{N}\}$, we can identify the training image with the highest content that can be projected onto this test pose. Subsequently, we select this training perspective as the projection target.
Following the approach outlined earlier, we can proceed to derive pseudo label for $\mathbf{P''}_n$.
We generate corresponding pseudo labels for all test poses $\{\mathbf{P''}_{1}, \mathbf{P''}_{2}, ..., \mathbf{P''}_{Q}\}$, where $Q$ represents the number of known test poses, 
then combine the training batch from these samples with the original training batch. The radiance field can be fine-tuned as the same optimization way in the previous section.

\begin{table*}[!t]
\centering
\resizebox{0.64\linewidth}{!}{
\begin{tabular}{c|c|c|ccc}
\hline
    Initial Pseudo &  ~Pseudo Label~ & Diverse Close-up   &\multicolumn{3}{c}{Novel View Synthesis} \\ 
    Label    &  Mask &   Perspective Generation  & ~PSNR $\uparrow$~   & ~~~~SSIM $\uparrow$~~~~ & ~LPIPS $\downarrow$ ~ \\ 
\hline
    -    &  -  & - &  14.48 & 0.657 & 0.552      \\
    $\surd$   & -  &  - & 16.61  & 0.683 & 0.548      \\
    $\surd$   & $\surd$  &  - &  18.17 & 0.700  & 0.520      \\
    $\surd$   & $\surd$  & $\surd$  & \textbf{18.92} & \textbf{0.710} & \textbf{0.515}    \\ 

\hline
\end{tabular}
}
\caption{Ablation studies of components in training on diverse close-up perspectives on our proposed dataset.
}
\label{tab:ablation_pseudolabel}
\end{table*}

\subsection{Dataset with Diverse Close-up Perspectives}
As shown in Fig.~\ref{fig:other_benchmark}, existing benchmarks for evaluating novel view synthesis methods typically only include test images within the training domain. However, since there is no available data to evaluate close-up view synthesis, we introduce a new dataset for evaluating the performance of current and future methods from close-up perspectives.

Our dataset comprises diverse scenes, each one is extracted as a subset of frames from a captured video, with 50 to 100 training images and 10 to 20 testing images for each scene, images are captured at a resolution of 960 $\times$ 540. As shown in Fig.~\ref{fig:benchmark}, our dataset has diverse scene types, contains both indoor and outdoor scenes. And the training images are closely resemble those in existing NeRF benchmarks like LLFF~\cite{mildenhall2019llff} and Tanks \& Temples~\cite{Knapitsch2017} (as in Fig.~\ref{fig:other_benchmark}), while the testing images concentrate on scenarios where camera poses brought much closer to objects and significantly differ from training views. 
In experiments section, we show the results of typical radiance field methods on our dataset, and the results reveal that our dataset effectively showcases the issues present in various methods within this specific setting. 

We follow the main evaluation metrics used in novel view synthesis methods to evaluate the performance on our dataset. Specifically, we use PSNR, SSIM~\cite{ssim} and LPIPS~\cite{lpips} to measure the quality of synthesized RGB novel views by comparing them with the ground truth images.

\section{Experiments}
\subsection{Implementation Details}
We conducted experiments following the implementation and settings of our baseline methods, NeRF~\cite{mildenhall2020nerf} and 2DGS~\cite{Huang2DGS2024}.
The experiments are conducted on NVIDIA 3090 GPUs and the Adam optimizer~\cite{kingma2014adam} is employed to optimize the radiance field. 
For both baseline methods, we first trained the radiance field following the implementation and settings of vanilla NeRF and 2DGS.
After that, for NeRF as baseline method, we load the weights of the pre-trained NeRF model and optimized for 10K iterations per scene with a ray batch of original training samples and generated samples set to 2048 (1024 for each).
For 2DGS as baseline method, we optimized the pre-trained 2DGS model for 500 iterations per scene with each iteration combine supervision from both training image and masked pseudo label.
For test-time fine-tuning, we applied our method on both NeRF and 2DGS. Specifically, we optimized the pre-trained 2DGS model with original training samples and generated samples together for iterations of 5 times for each test view. And for NeRF, we optimized the pre-trained model for 200 iterations on test poses, with a batch size of 1024 for each samples.

\subsection{Ablation Studies}
In this section, we perform comprehensive ablation studies to evaluate the effectiveness of our method, including the effectiveness of training on diverse close-up perspectives and the effectiveness of test-time fine-tuning.

\subsubsection{The Effectiveness of Components in Training on Diverse Close-up Perspectives.}
For the initial pseudo labels in Tab.~\ref{tab:ablation_pseudolabel}, we directly use the warped result $\mathbf{I'}_{n}$ to generate batch samples in the fine-tuning process.
For pseudo label mask, we use mask $\mathbf{M}_{n}$ to select out the reliable areas in $\mathbf{I'}_{n}$ for the fine-tuning process. 
For the fine-tuning without diverse close-up perspective generation, we manually defined four anchor points for each training image and fixed the distance to create four consistent generated poses for each training viewpoint.
As shown in Tab.~\ref{tab:ablation_pseudolabel}, directly employing the warped results as pseudo labels for fine-tuning brings a relatively limited improvement over the baseline, increasing from 14.48 to 16.61 in PSNR. This is because the warped pseudo labels contain inaccuracies, and using them directly for training partially address the issue of artifacts, but simultaneously introduce new errors.
Therefore, applying pseudo label masks to filter out erroneous regions can lead to substantial improvements in results, highlighting the crucial function of pseudo label masks in enhancing fine-tuning effectiveness. As we can observe from Tab.~\ref{tab:ablation_pseudolabel}, applying pseudo label masks improves PSNR from 16.61 to 18.17.
The effectiveness of randomly generating diverse close-up perspectives is evaluated in the final row of Tab.~\ref{tab:ablation_pseudolabel}, it further improves PSNR from 18.17 to 18.92 by maintaining pose diversity, which results in a variety of rays for training.

\begin{table}[!b]
\begin{center}
{
\resizebox{1\linewidth}{!}{
\begin{tabular}{l|ccc}
\hline
Method   & ~PSNR$\uparrow$~ & ~SSIM$\uparrow$~ & ~LPIPS$\downarrow$~ \\
\hline
NeRF~\cite{mildenhall2020nerf} &  $14.48$  &  $0.657$  &  $0.552$ \\
2DGS~\cite{Huang2DGS2024} &  $19.99$  &  $0.768$  &  $0.386$ \\
Ours-NeRF &  $18.89~(\mathbf{+4.41})$  &  $0.708$  &  $0.520$ \\
Ours-2DGS &  $\mathbf{20.95~(+0.96)}$  &  $\mathbf{0.785}$  &  $\mathbf{0.383}$ \\
\hline
\end{tabular}
}
\caption{Ablation studies of test-time fine-tuning. We directly use test poses of each scene in our dataset to fine-tune the radiance field, and also test on those poses. 
}
\label{tab:cmp_testtime}
}
\end{center}
\end{table}

\begin{figure*}[!t]
\centering
\includegraphics[width=0.98\linewidth]{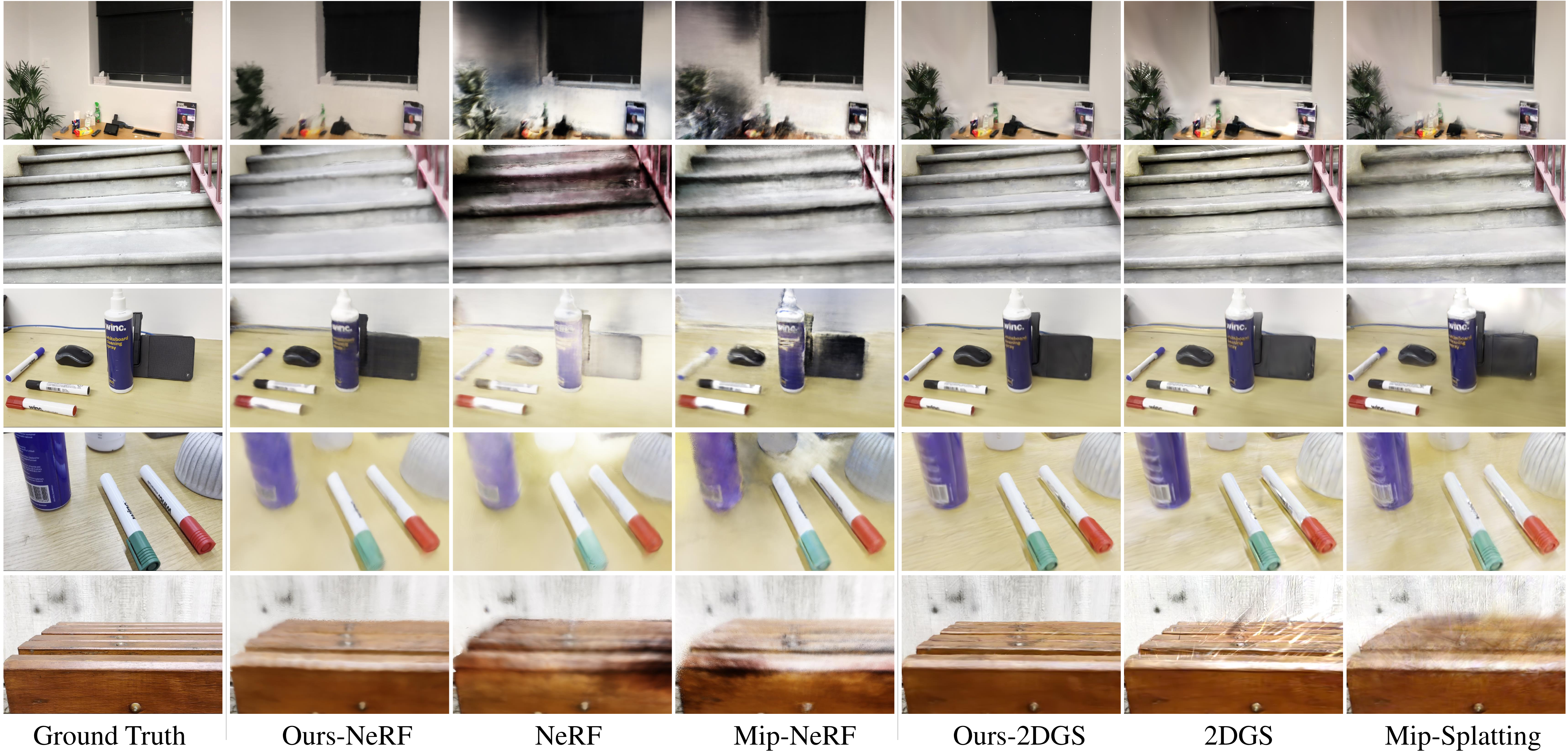}
\caption{Quantitative comparisons with other methods. We visualize the synthesized images from our method, and compare them with Mip-NeRF, Mip-Splatting and the baseline methods.
} 
\label{fig:comparison}
\end{figure*}

\subsubsection{The Effectiveness of Test-time Fine-tuning.}
If users know which close-up perspectives they need to enhance performance for, we can directly fine-tune an existing model by generating and applying pseudo-labels for those specific perspectives. To demonstrate this, we conducted comparison experiments using NeRF and 2DGS.
Regarding processing speed, our experiments revealed that fine-tuning a NeRF model using the proposed test-time fine-tuning method requires only 200 iterations for around 10 testing views, while fine-tuning a 2DGS model is also very fast, requiring just 5 iterations per testing view.
In terms of accuracy, as shown in Tab.~\ref{tab:cmp_testtime}, our test-time fine-tuning method on 2DGS achieves a PSNR of 20.95, compared to the baseline 2DGS PSNR of 19.99. This approach significantly speeds up processing while achieving outstanding performance, even comparable to our general training method.
The primary reason for this is that general close-up perspectives training aims to cover diverse perspectives, requiring more iterations for learning. 
We argue that the proposed test-time fine-tuning does not conflict with the general training method, as it is designed for specific cases. For a general scenario where testing poses are unknown before training, the proposed general training method remains applicable.

\begin{table}[!t]
\begin{center}
{
\resizebox{1\linewidth}{!}{
\begin{tabular}{l|ccc}
\hline
Method   & ~PSNR$\uparrow$~ & ~SSIM$\uparrow$~ & ~LPIPS$\downarrow$~ \\
\hline

Instant-NGP~\cite{mueller2022instant} &  $14.23$  &  $0.615$  &  $0.616$ \\
TensoRF~\cite{chen2022tensorf} &  $14.59$  &  $0.651$  &  $0.613$ \\
NeRF~\cite{mildenhall2020nerf} &  $14.48$  &  $0.657$  &  $0.552$ \\
Zip-NeRF~\cite{barron2023zipnerf} &  $15.31$  &  $0.637$  &  $0.510$ \\
DS-NeRF~\cite{deng2022depth} &  $16.48$  &  $0.676$  &  $0.558$ \\
Mip-NeRF~\cite{barron2021mip} &  $17.05$  &  $0.665$  &  $0.554$ \\
3DGS~\cite{kerbl3Dgaussians} &  $19.16$  &  $0.747$  &  $0.430$ \\
Mip-Splatting~\cite{Yu2024MipSplatting} &  $19.55$  &  $0.765$  &  $0.407$ \\
2DGS~\cite{Huang2DGS2024} &  $19.99$  &  $0.768$  &  $0.386$ \\

\hline
Ours-NeRF &  $\mathbf{18.92}$  &  $\mathbf{0.710}$  &  $\mathbf{0.515}$ \\
Ours-2DGS &  $\mathbf{20.88}$  &  $\mathbf{0.784}$  &  $\mathbf{0.383}$ \\

\hline
\end{tabular}
}
\caption{Qualitative comparisons of diverse close-up fine-tuning with other methods.}
\label{tab:cmp_diverse}
}
\end{center}
\end{table}


\subsection{Comparisons with Other Methods} 
In this section, detailed comparisons between our general fine-tuning method and other approaches are presented. 

\subsubsection{Comparisons of Numerical Accuracy.}
As shown in Tab.~\ref{tab:cmp_diverse}, we can observe that our method significantly outperforms other methods in terms of all the metrics.
For NeRF-based methods,
our method achieves PSNR of 18.92, SSIM of 0.710 and LPIPS of 0.515, demonstrates a significant performance improvement in PSNR (+4.44) compared to our baseline, NeRF. 
Our method also shows a substantial improvement compared to Instant-NGP, TensoRF and Zip-NeRF.
Mip-NeRF demonstrates some robustness when bringing the camera closer compared to NeRF. 
However, in comparison to our method, it shows difference in PSNR (-1.87), SSIM (-4.5\%) and LPIPS. DS-NeRF benefits from its utilization of depth constraints, shows good performance in reconstructing geometric information and exhibits robustness when close-up the viewpoint, but still has significant performance gap from our method.
For Gaussian Splatting methods,
our method achieves PSNR of 20.88, SSIM of 0.784 and LPIPS of 0.383, effectively improving the baseline method 2DGS across all metrics, significantly outperforming 3DGS in PSNR (+1.72), SSIM (+3.7\%) and LPIPS (-4.7\%).
Mip-Splatting was proposed to address alias-free rendering, however, it fails to handle the close-up case, achieving only a PSNR of 19.55 (-1.33).
These comparisons clearly highlight the remarkable improvement of our method when close-up observations are required.

\subsubsection{Comparisons of Visualized Results.}
The visualization results on our benchmark in Fig~\ref{fig:comparison} clearly demonstrate the problem that other conventional methods face under the settings proposed in this paper. 
Especially in the last row of comparisons, each method's respective issues are most pronounced.
We can observed that the results of NeRF are affected by the issue of artifacts, as we analyzed in our paper. Mip-NeRF, 2DGS and Mip-Splatting also shows similar issues, demonstrating that these challenges are prevalent across radiance field methods.
Our method clearly addresses the issue of artifacts compared to our baseline methods. The visualization results demonstrate a notable improvement in the rendering quality, indicating a significant resolution to the problems outlined in our paper.

\section{Conclusion} 
This work dig into the complexities of generating reliable images from viewpoints that diverge considerably from the training data, especially for close-up perspectives. We propose a novel learning strategy utilizes pseudo-labels derived from the available training data to offer targeted guidance for various close-up viewpoints to tackle this challenge. We also developed a dataset tailored to evaluate the synthesis of close-up views, addressing the lack of a benchmark for assessing the effectiveness of existing and upcoming methods in this particular task.


\bibliography{aaai25}

\end{document}